# Particle Swarm Optimization Based Diophantine Equation Solver

**Siby Abraham***
Department of Mathematics & Statistics, G.N. Khalsa College,
University of Mumbai, Mumbai-400019, India
E-mail: sibyam@gmail.com
*Corresponding author

**Sugata Sanyal**
School of Technology & Computer Science
Tata Institute of Fundamental Research, Mumbai, India
Homi Bhabha Road, Mumbai-400005, India
E-mail: sanyal@tifr.res.in

**Mukund Sanglikar**
Department of Mathematics, Mithibai College, Vile Parle (W),
University of Mumbai, Mumbai-400056, India.
Email: masanglikar@rediffmail.com.

**Abstract:** The paper introduces particle swarm optimization as a viable strategy to find numerical solution of Diophantine equation, for which there exists no general method of finding solutions. The proposed methodology uses a population of integer particles. The candidate solutions in the feasible space are optimised to have better positions through particle best and global best positions. The methodology, which follows fully connected neighbourhood topology, can offer many solutions of such equations.

**Keywords:** Diophantine equation, particle swarm optimisation, fitness function, position, velocity, learning factor, socio-cognitive coefficients, Fermat's equation, elliptic curve.

**Biographical notes:** Siby Abraham is an Associate Professor and Head, Department of Mathematics & Statistics at Guru Nanak Khalsa College, University of Mumbai. He is also a visiting faculty to Department of Computer Science, University of Mumbai, India. He received his PhD in Computer Science from University of Mumbai and MSc in Mathematics from Cochin University of Science & Technology, India. His current research interests include computational intelligence, nature and bio inspired computing.

Sugata Sanyal is a Professor at the School of Technology & Comp Sc at Tata Institute of Fundamental Research (TIFR), India. He received his PhD from TIFR under Mumbai University, M Tech from IIT, Kharagpur and B E from Jadavpur University, India. He was a visiting professor in the University of Cincinnati, USA. He is in the editorial board of three international journals. His current research interests include security in wireless and mobile ad hoc networks, distributed processing scheduling techniques and bio inspired computing.

Mukund Sanglikar is presently with the Department of Mathematics, Mithibai College, Mumbai, India. He is former Professor and Head of the Department of Computer Science, University of Mumbai, India. He was also the Chairman, Board of Studies in Comp Sc & IT and a member of Academic Council, University of Mumbai. He received his PhD in Computer Science and MSc in Mathematics from University of Mumbai, India. His current research interests include soft computing, networking, software engineering and mobile databases.

## 1 INTRODUCTION

A Diophantine Equation (Cohen 2007) (Rossen 1987) (Zuckerman 1980) is a polynomial equation, given by

$$f(a_1,a_2,\ldots,a_n,x_1,x_2,\ldots,x_n) = N \quad (1)$$

where $a_i$ and N are integers. Diophantine equations obtained its name from third century BC Alexandrian Mathematician Diophantus, who studied these equations in detail (Bashmakova 1997). In his work 'Arithmetic', a treatise of thirteen books, he asked for solutions for about 150 algebraic problems. These problems are now collectively referred as Diophantine equations.

Diophantine equations and its particular cases have always been of great interest to Mathematicians (Bag 1979). One of the simplest forms of such equations is given by

$$ax_1 + bx_2 = c \quad (2)$$

If c is the greatest common divisor of 'a' and 'b' (i.e. c = (a, b)), then this equation has infinitely many solutions, which can be obtained by extended Euclidian Algorithm.

The equations given by

$$x_1^n + x_2^n = x_3^n \quad (3)$$

are of great significance. For n=1, the equation (3) has infinitely many solutions. For n=2 also, the equation provides infinitely many solutions. Such solutions, which are called Pythagorean triplets, form sides of a right angled triangle.

The French Mathematician Pierre de Fermat's name is the most relevant in the discussion of Equation (3). Fermat, who died in 1665, had the habit of writing small notes on the margins of the book he read. In one of such notes made on the Latin translation of Diophantus' 'Arithmetic' by Backet, he wrote that the equation (3) has no solutions for n>2. The notes appeared alongside problem 8 in Book II of 'Arithmetic' says "…given a number which is square, write it as a sum of two other squares". Fermat's note further stated that "on the other hand, it is impossible for a cube to be written as a sum of two cubes or a fourth power to be written as sum of two fourth powers or, in general, for any number which is a power greater than the second to be written as a sum of two like powers? I have a truly marvelous demonstration of this proposition which this margin is too narrow to contain" (Shirali & Yogananda 2003). Since then, this conjecture is known as Fermat's Last Theorem (FLT), though there was no general proof of it irrespective of the number of attempts to find one. Table 1 shows the important attempts by Mathematicians to prove FLT for different cases (Edwards 1977). Finally, in 1994, the British Mathematician Andrew Wiles presented a sophisticated proof using elliptic curves for FLT and ended the conjecture. Thus Fermat's Last Theorem became a theorem at last! (Shirali & Yogananda 2003)

An elliptic curve is a Diophantine equation (Stroeker and Tzanakis 1994)(Poonan 2000) of the form

$$y^2 = x^3 + ax + b \quad (4)$$

where a and b are rational and the cubic $x^3 + ax + b$ has distinct roots. Elliptic curves (Koblitz 1984) form a highly advanced structure because of its mathematical rigor.

| Year | Mathematician | Case proved |
|---|---|---|
| 1640 | Fermat | n=4 |
| 1770 | Euler | n=3 |
| 1823 | Sophie Germaine | n=p, a prime known as Sophie Germaine prime where 2p+1 is also prime |
| 1825 | Dirichlet, Legendre | n=5 |
| 1832 | Dirichlet | n=14 |
| 1839 | Lame | n=7 |
| 1847 | Kummer | n=p, where p is a regular prime |

**Table 1:** *Status of FLT for different cases*

If P and Q are any two points on an elliptic curve, then we can uniquely describe a third point which is the intersection of the curve with the line through P and Q. The work of Poincare showed that the points of E (real not just rational) form an abelian group under a specific operation known as chord-and-tangent addition with the 'point at infinity' O as the identity element. Such Mathematical sophistication makes elliptic curves a participant in high level applications. For example, Elliptic curve based Public Key Cryptography (Lin CH et al 1995)(Laih CS et al 1997) is quite effective as discrete logarithmic problem on elliptic curve is tougher than usual discrete logarithmic problems. Hence, such systems require only shorter key size to have comparable security offered by other public key cryptosystems.

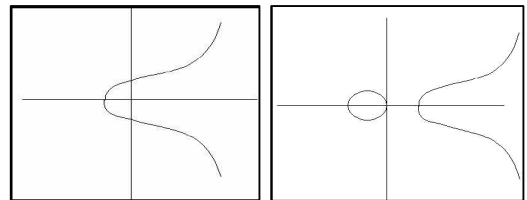

Figure 1 **Elliptic curves**

Though there are different types of well known and highly relevant Diophantine equations, the solutions of which have always been surrounded by an air of enigma. This is because a Diophantine Equation can have no nontrivial solution, a finite number of solutions or an infinite number of solutions. David Hilbert should be given the credit for giving a direction to the interest on Diophantine equations and its solutions. In 2000, at the second international conference of Mathematicians (Hilbert 1902), he asked "Given a Diophantine equation with any number of unknowns and with rational integer coefficients: devise a process, which could determine by a finite number of operations whether the equation is solvable in rational integers" as tenth of his famous twenty three problems. Since then, the problem is known as Hilbert's tenth problem (Borowski et al 1991).

There have been many attempts to solve Hilbert's tenth problem. Davis et al (1982) showed that an algorithm to determine the solvability of all exponential Diophantine equations is impossible. Matiyasevich (1993) extended that work by showing that there is no algorithm for determining whether an arbitrary Diophantine equation has integral solutions. This helped in ending the search of centuries for finding a general method to solve a Diophantine equation.

This has not dampened the importance and interest on Diophantine Equations and its solutions as newer and modern application areas were added. These include Public key cryptosystems (Laih CS 1997) (Lin CH 1995), Data dependency in Super Computers (Tzen and Ni 1993) (Zhiyu et al 1989), Integer factorization, Algebraic curves (Ponnen 2000), Projective curves (Brown & Myres 2002) (Stroeker & Tzanakis 1994), Computable economics (Velu 2003) and Theoretical Computer Science (Ibarra 2004) (Guarari 1982).

In this context, finding numerical solution to Diophantine equations becomes important. However, this turns out to be a tough problem to deal with primarily because of the fact that the number of possible solutions encountered is very huge. This means, following an exhaustive, gradual and incremental method invite the definite risk of computational complexity.

This paper introduces Particle Swarm optimization (Eberhart et al 1995) (Kennedy et al 1995) as a meta-heuristic search technique to find numerical solutions of such equations. It shows how the socio psychological behavior of custom made n-dimensional integer particles effective in maneuvering the search space in finding a solution to such equations. The paper discusses the procedure which are validated using a class of Diophantine equations given by

$a_1 \cdot x_1^{p1} + a_2 \cdot x_2^{p2} + \ldots\ldots + a_n \cdot x_n^{pn} = N$ (5)

Other class of equations can also be tackled similar way.

The paper is organized as follows: Section 2 gives a brief survey and discussion of related works. Section 3 explains the methodology used. Section 4 presents the experimental results and section 5 deals with conclusion.

## 2  BRIEF SURVEY AND DISCUSSION

Since there exists no general method to find solution of a Diophantine Equation, there have been few attempts to find numerical solutions of it as the next possible way. However, this turns out to be a tough problem to deal with as there are $N^n$ possible in the search space of (5). The search strategies offered by Artificial Intelligence (AI) (Russell and Norwig 2003) (Luger 2006) could be possible alternatives because of its flexibility to move around in the search space. Though Breadth first search is guaranteed to find an optimal solution, if it exists, the time and space complexities of order $O(b^{d+1})$ with branching factor 'b' and depth 'd', discourage us to attempt it for larger equations. Depth first search might blindly follow a branch of the search tree without returning a solution. The other uninformed search strategies like Depth Limited, Iterative Deepening and Bi-directional also can come with similar handicaps. The informed search technique Hill Climbing can get trapped in local optimum points from where it finds it difficult to get out. A* search was used by Abraham and Sanglikar (2009) and found that the system runs out of space very fast.

There have been some attempts to apply soft computing techniques to find numerical solution of (5). Abraham and Sanglikar (2001) tried to find numerical solutions to such Diophantine equations by applying genetic operators-mutation and crossover Michalewich (1992). Though the methodology could find solutions, it was not fully random in nature and seemed more like a steepest ascent hill climbing rather than a genetic algorithm. Hsiung and Mathews (1999) tried to illustrate the basic concepts of a genetic algorithm using first-degree linear Diophantine equation given by a + 2b + 3c+ 4d = 30. Literature also talks about an application of higher order Hopfield neural network to find solution of Diophantine equation (Joya et al 1991). Abraham and Sanglikar (2007 a) explains the process of avoiding premature converging points using 'Host Parasite Co-evolution' (Hills 1992) (Paredis 1996) (Wiegand 2003) in a typical GA. Later they used a method involving evolutionary and co-evolutionary computing (Rosin and Belew 1997) techniques (Abraham & Sanglikar 2007 b) to find numerical solutions to such equations. They also used (Abraham & Sanglikar 2008) Simulated Annealing as a viable probabilistic search strategy for tackling the problem of finding numerical solution. These methods, though effective to a certain extent for smaller equations, are not good enough to deal with the complexities of Diophantine equations.

## 3  SWARM-DOES METHODOLOGY

Though PSO algorithm is designed for real-value problems (Shi 2004), there have been few attempts to apply them to binary-value problems (Kennedy and Eberhart 1997) (Agrafiotis and Cedeno 2002). This paper is an attempt to apply the PSO methodology to integer value problems. More importantly, this tries to apply the algorithm to a Mathematical problem, an area which appears not so often in the PSO literature.

The system developed to find numerical solution of Diophantine equations using particle swarm optimization, is called Particle Swarm Optimization based Diophantine Equation Solver (SWARM-DOES). SWARM-DOES attempts to apply PSO on the integer value problem of finding integer solutions of (5). The particles are created as integer particles to facilitate integer solutions. The dimension of the particle depends upon the number of variables of the equation under study. If an equation has n variables, the dimension of the particle would also be 'n'. As in the other PSO systems, each particle is identified with two factors: the velocity and position. The velocity factor shows the measure of the movement of the particle and the position conveys the current status of the particle. Each of the particle tries to change its velocity and hence the position. They move in the multidimensional space in a collective fashion. The objective of the movement of these particles is to find the best position a particle can have in this multidimensional search space. The strategy involves mimicking the best position it had and the best position other particles have had till then.

### 3.1 Initial population

The procedure of finding numerical solution to (5) starts with a population of random particles or swarms of fixed size. The particles are constructed as integer particles based on probable values of variables appearing in the equation (5). The construction of these particles is facilitated by incorporating knowledge of the domain and the constraints the possible values face as solutions in the problem.

A possible solution of equation (5) can be $(s_1, s_2, \ldots, s_n)$ where each $s_i$ is an integer which lies between the numbers 1 and the $p_r^{th}$ power root of N, where $p_r$ is the minimum value of $p_1, p_2, \ldots$ and $p_n$. Hence a particle in the initial population is taken as an n-dimensional vector $(d_1, d_2, \ldots, d_n)$ where each of the coordinate $d_i$ takes a random integer value between 1 and integer part of $N^{1/p_r}$. Thus, the initial population is a random collection of fixed number of integer particles which occupy random positions at the search space of candidate solutions of a Diophantine equation. Initially, each of the particles is given velocities zero. The population size can be chosen as any relevant value.

### 3.2 Feasible space

SWARM-DOES procedure separates the positions a particle occupies into two spaces: the feasible space and the non-feasible space. The feasible space consists of all positions, which a particle can fly legally without violating the constraints of the problem. The non-feasible space consists of all other positions of the particle. The feasible space X of the equation (5) is defined as

$X = \{(d_1, d_2, \ldots, d_n): d_i = 1, 2, 3, \ldots N^{1/p_r}\}$ (6)

where $p_r$ is given by

$p_r = \text{minimum } \{p_1, p_2, \ldots, p_n\}$ (7)

Here $p_1, p_2, \ldots, p_n$ are the powers of the Diophantine equation (5). This selection is based on the domain knowledge that the equation (5) cannot have a solution whose power is less than the $p_r^{th}$ power root of N. Thus, the problem of finding numerical solution of a Diophantine equation is a constrained problem of finding numerical solution of the equation (5) by searching within the feasible space X. The proper selection of feasible space helps in identifying to check whether a candidate solution satisfies the constraints of the given problem. As shown in the next section, the identification of candidates belonging to feasible solution is made possible by introducing a domain specific fitness function. The particle with a valid and acceptable fitness function value only are allowed to be in the feasible space. The candidates with invalid fitness function are directed to belong to the infeasible space.

### 3.3 Fitness function

Fitness function value of a particle gives the effectiveness of the particle in the search space. The fitness function of a particle in SWARM DOES is defined as

fitness $= \text{Abs}(N - (a_1 * x_1^{p_1} + a_2 * x_2^{p_2} + \ldots + a_n * x_n^{p_n}))$ (8)

The value of the fitness function indicates the distance between the current position of the particle and its solution. If fitness=0 for a particle, then that position of the particle is taken as a solution. At each iteration, the attempt is to reduce this distance. Thus, the procedure becomes a minimization process in which each of the particles tries to reduce the distance between its present status and the solution of the equation whose fitness function value is given to be zero.

### 3.4 Selection of Pbest

The movement of a particle in the search space is guided by the previous best experience of the particle and the best experience encountered by all the particles till then. The best position of a particle faced till then is called the 'pbest' of that particle. The 'pbest' acts as the memory of the particle. Based on the best position encountered by that particle during the exploration process, the particle tries to mimic that best experience by duplicating the same. Thus, each particle memorizes one best experience it encountered till then and tries to optimize the current position by comparing the best experience it had. The fitness function value of a particle acts as a representative value in the selection of 'pbest' of the particle at that iteration. A particle with the lowest fitness function value is taken as the 'pbest' of the particle. Here the particle remembers only the states which are feasible and discard the particles in the infeasible space. It is possible that 'pbest' position of a particle may survive for a longer time in the search procedure. Better the value of the fitness of the 'pbest', higher the chance for the particle to survive for a long time. In a highly matured run of the process, 'pbest' values do not survive for a long time. The dynamic movement of 'pbest' values shows the effectiveness of convergence of the procedure. At the beginning of the process, 'pbest' of each of the particle is initialized to the initial fitness value of that particle.

### 3.5 Selection of gbest

The movement of a particle in the multidimensional search space of a Diophantine equation also depends on the best position encountered by all particles in the population till then. This way, the flow of a particle is influenced by the social environment in which it flows (Hu et al 2004). Each of the particles tries to improve its position by comparing the best position encountered by any particle till then during the entire exploration process. This best position of all the particles in the entire the search process is called the 'gbest'. Initially, the particle having the best fitness function value is taken as the 'gbest'. This is changed dynamically as the process continues. This is made possible by having a reference to the stored value of the 'gbest'. This reference is used to update the position and velocity of each of the particle in the population. At any instant, the particle changes its velocity by comparing the same with that of the 'gbest' particle. Hence the 'gbest' particle helps in updating the position of each of the particle in the whole search process.

### 3.6 Update of velocity

The movement of a particle $x_t$ at time t to that at time t+1 is facilitated by updating the velocity and hence the position of the particle. The velocity $v_t$ of the particle $x_t$ at time t is updated using the formula

$$V_{t+1} = c_1 v_t + c_2.rand1().(x_t - pbest) + c_3.rand2().(x_t - gbest) \quad (9)$$

where rand1() and rand2() are two random numbers between 0 and 1. Since $x_t = (d_1, d_2, ......, d_n)$, the updating formula for the velocity for each coordinate of a particle is given by

$$V_{t+1}(d_i) = c_1 v_t(d_i) + c_2.rand1().(x_t(d_i) - pbestd_i) + c_3.rand2().(x_t(d_i) - gbestd_i) \quad (10)$$

The component $(x_t(d_i) - pbestd_i)$ gives the difference of the corresponding coordinates of the current position and that of the 'pbest' position of the particle. $c_2$ is a constant, called the cognitive coefficient, denotes the trusts the particle has on its own previous experience. Thus, the factor rand1()$(x_t(d_i) - pbestd_i)$ identifies the amount of change in velocity of a particle due to its own previous experience.

The component $(x_t(d_i) - gbestd_i)$ gives the difference of the corresponding coordinates of the current position of the particle and that of the 'gbest' position. The social coefficient $c_3$ quantifies the trusts the particle has on its own neighbors. Hence, the factor $c_3.rand2().(x_t(d_i) - gbestd_i)$ represents the amount of change in velocity of a particle due to its neighbors.

### 3.7 Update of position

A particle changes its position using its updated velocity. The following formula is used for this purpose:

$$x_{t+1} = x_t + v_{t+1} \quad (11)$$

Thus, the coordinates of the new position obtained by a particle is given the formula

$$x_{t+1}(d_i) = x_t(d_i) + v_{t+1}(d_i) \quad (12)$$

Since the paper is interested only in non-negative solutions of the given Diophantine equation, those values of $X_{t+1}(d_i)$ which belong to the non-feasible space are discarded. The occurrence of positions belonging to non-feasible space demands an automatic revision. This is facilitated by a unique strategy in SWARM-DOES.

According to this, any particle which risks to flow into the non-feasible space is brought back to the feasible space.

It has been observed that a particle in the procedure can fall into the non-feasible space in two cases:
Case 1: When coordinate of any particle $d_i < 0$;
Case 2: $d_i >$ particleRange, where

$$particleRange = N^{1/pr} \quad (13)$$

In case 1, the particle is made to flow into the feasible space by suitably redefining $d_i$ as

$$d_i = (-1) * d_i \quad (14)$$

In the case 2 also, the particle is forced to fly back to the feasible space but using the instruction:

$$d_i = d_i \% \text{ particleRange} \quad (15)$$

In either case, a particle is not allowed to fly into the non-feasible space. This strategy of not allowing a particle the risk of flying into the non-feasible space by forcefully diverting its route is quite effective in managing the constraints of the given Diophantine equation.

### 3.8 Neighbourhood topology

SWARM-DOES follow the fully connected neighborhood topology (Sierra and Coello 2006), which is also called star topology (Engelbrecht 2002), against local topologies. As per this, each integer particle flies through the search space by dynamically adjusting its position with respect to 'pbest' and 'gbest' (Kennedy 1999) (Kennedy et al 2001) (Kennedy and Mendes 2002) (Shi 2004). Thus, the graph obtained to show the connections between the particles is a fully connected graph as shown in figure 2. The circle shows the particles and the line segments show the connection. In such a topology, all members of the swarm are connected to one another. The adoption of this fully connected topology against other popular topologies is deliberate. We felt that the other topologies, by the virtue of guiding the search process by focusing only on the best position encountered.

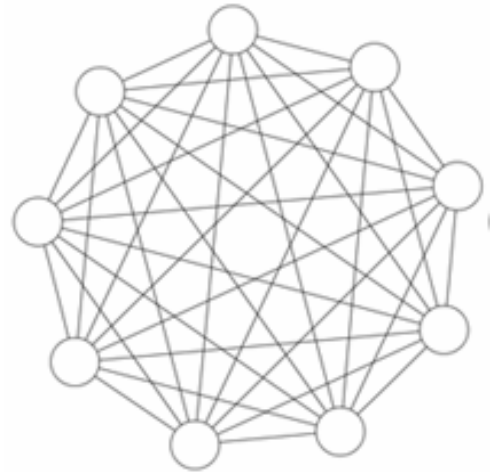

**Figure 2** Fully connected graph

by the particles in a neighborhood topology, could converge only slowly. On the other hand, the global version can provide a faster convergence by focusing on the best positions encountered by all particles by taking the whole population as its neighbors. This is because, all the particles receive information about the best position of the entire

swarm simultaneously and are better equipped to incorporate this information

Having the complete population of particles as neighbors brings a lot of convenience and effectiveness to the procedure. The best experience encountered by any particle can be shared by the other particles in this mechanism. This makes the system much more responsive than other local level topologies which focus only on a limited number of neighbors in the topological vicinity of a particle.

### 3.9 Learning factors

As the equation (10) shows, SWARM-DOES incorporate the velocity changes of a particle in three parts. As is the popular convention, we call them momentum part, cognitive part and social part. The momentum part corresponds to the factor $c_1 v_t(d_i)$ of the equation (10). The constant $c_1$, called the inertia weight, balances the global exploration and local exploitation of the search. It has been experimentally shown that large inertia weight increases the global search while the small value supports the local search (Shi 2004).

In equation (10), the factor $c_2(x_t(d_i)–pbestd_i)$ conveys the coordinate wise effect of the current position of the particle and the best position encountered by the particle till then. The particle tries to optimize its performance by comparing its best performance. This factor is enhanced using the cognitive factor $c_2$. On psychological terms, $c_2$ shows the tendency of the particle to duplicate the past best behavior.

The factor $c_3.rand2().(x_t(d_i)- gbestd_i)$ indicates the quantity by which the velocity of a particle can change based on the best performance of all particles till then. The constant $c_3$, called the social coefficient represent the confidence with which a particle can follow the success of other particles.

Since the effectiveness of the search process to a great extend depends on the choice of the values of the values of the constants $c_1$, $c_2$ and $c_3$, we have taken care to select the optimum values for these. After experimental results we finalized the value of $c_1$ as 1 and the values of $c_2$ and $c_3$ as 2 each. Though there have been many varied forms of inertia weight (Shi and Eberhart 1998a, 1999) (Shi and Eberhart 2001a, 2001b) (Ratnaweera et al 2004), SWARMDOES kept the value of $c_1$ as 1 deliberately. The possible effect of other values of $c_1$ is brought more effectively by incorporating a clamp on the possible values of velocity as the section 3.10 demonstrates.

### 3.11 Maximum velocity

In order to restrict the change in velocity of a particle within an acceptable range, a variable called $V_{max}$ is used. Using this concept, the equation (10) is not given a free hand to change the value of the velocity of a particle. However, the values of the velocity are allowed to change only within an acceptable range. It is facilitated using a specially defined methodology. As per this, based on the value of 'particleRange' defined in equation (13), we define a new variable maxRange as given below

$$\text{maxRange} = \text{minimum} \{\text{particleRange}, 5\} \quad (16)$$

Then, the velocity coordinates are allowed to change only within the ± maxRange. That means, the coordinates of the velocity are allowed to change only within the region given by

$$(-v_{max}, v_{max}) = (-\text{maxRange}, \text{maxRange}) \quad (17)$$

In cases, where the velocity values go outsides this range, the procedure brings them back to the feasible range by using the mathematical equation

$$v_{t+1}(d_i) = v_{t+1}(d_i) \ \% \ \text{maxRange} \quad (18)$$

Having the restriction on the possible values of velocity allows us to have a much thorough local search of the search space.

### 3.11 Sequence of solutions

The SWARM-DOES procedure is run many times until get a particle whose fitness value is zero or termination condition is met. After getting one solution, the procedure offers to continue further to get as many solutions of the equation as possible. This requires some fine tuning in the procedure explained until now. During experiment, it is observed that after finding the first solution, the usual procedure of updating 'pbest' and 'gbest' result in the repetition of the same solution again and again. The system overcomes this problem by replacing the particle, which gives the solution, by a randomly generated new particle with velocity assigned as zero. This would enable the optimization to search another untried and untested domain. Leaving an optimised region and charting an entirely new untried and new region helps the procedure to find as many numerical solutions as possible for a given Diophantine equation.

### 3.12 Termination condition

The procedure is terminated when the number of generations specified is completed. Or, alternatively, the procedure halts when the system offers one solution asking for the option of continuing the search to produce newer and fresher solutions of the given equation. Here, one can terminate the procedure if satisfied with one solution. The option of offering newer solutions helps to find as many numerical solutions as possible for a given Diophantine equation. This is quite significant, in the light of the fact, that there exists no general method even to find a single solution. The system experienced the running of system up to more than fifty thousand generations or iterations.

## 4    EVALUATION AND EXPERIMENTAL RESULTS

The SWARM-DOES strategy was validated by testing the system with different types of equations. The different classes of Diophantine equations given in (5) were tried by supplying different number of variables and varied values of degrees. These results also reveal some important characteristics of swarm intelligence while finding the solution of the given Diophantine equation. The important features the results demonstrate include convergence properties of the 'pbest' and 'gbest' particles, directed random movement of the procedure during the search and the random update of the values of velocity among others.

### 4.1 Results on equations with varying degrees

Table 2 demonstrates results obtained for Diophantine equations with different degrees. The table contains equations with powers changed from 2 to 15. These equations are significant as the powers increase, the value of 'N' also increases significantly. The table demonstrate that the system could give solutions quite faster, irrespective of higher values of powers. It also shows, as discussed in detail in section 4.3, that the system does not require larger number of iterations even when the values of N are quite high. This is because the feasible space involved is much less than the space of all possible values. The system could provide solutions much faster because of the number of possible solutions it has to search are much lesser as compared to other equations with smaller values of powers. Higher the values of powers, better the chance of getting the equations faster. These results demonstrate the effectiveness and the convenience the system to find solutions.

| Sr. No | Diophantine Equation | Degree of Eq. | Solution Found |
|---|---|---|---|
| 1 | $x_1^2 + x_2^2 = 625$ | 2 | 24, 7 |
| 2 | $x_1^3 + x_2^3 = 1008$ | 3 | 2, 10 |
| 3 | $x_1^4 + x_2^4 = 1921$ | 4 | 6, 5 |
| 4 | $x_1^5 + x_2^5 = 19932$ | 5 | 5, 7 |
| 5 | $x_1^6 + x_2^6 = 47385$ | 6 | 6, 3 |
| 6 | $x_1^7 + x_2^7 = 4799353$ | 7 | 9, 4 |
| 7 | $x_1^8 + x_2^8 = 16777472$ | 8 | 2, 8 |
| 8 | $x_1^9 + x_2^9 = 1000019683$ | 9 | 3, 10 |
| 9 | $x_1^{10} + x_2^{10} = 1356217073$ | 10 | 7, 8 |
| 10 | $x_1^{11} + x_2^{11} = 411625181$ | 11 | 6, 5 |
| 11 | $x_1^{12} + x_2^{12} = 244144721$ | 12 | 5, 2 |
| 12 | $x_1^{13} + x_2^{13} = 1222297448$ | 13 | 3, 5 |
| 13 | $x_1^{14} + x_2^{14} = 268451840$ | 14 | 2, 4 |
| 14 | $x_1^{15} + x_2^{15} = 1088090731$ | 15 | 4, 3 |

**Table 2:** *Results on equations with varying degrees*

| Sr. No | Diophantine Equation | No. of variables | Solution Found |
|---|---|---|---|
| 1 | $x_1^2 + x_2^2 = 149$ | 2 | 7, 10 |
| 2 | $x_1^2 + x_2^2 + x_3^2 = 244$ | 3 | 6, 12, 8 |
| 3 | $x_1^2 + x_2^2 + … + x_4^2 = 295$ | 4 | 2, 11, 11 |
| 4 | $x_1^2 + x_2^2 + …. + x_5^2 = 325$ | 5 | 14, 10, 4, 2, 3 |
| 5 | $x_1^2 + x_2^2 + …. + x_6^2 = 420$ | 6 | 3, 16, 7, 9, 3, 4 |
| 6 | $x_1^2 + x_2^2 + …. + x_7^2 = 450$ | 7 | 12, 6, 6, 7, 4, 5, 12 |
| 7 | $x_1^2 + x_2^2 + …. + x_8^2 = 590$ | 8 | 7, 9, 7, 12, 13, 8, 5, 3 |
| 8 | $x_1^2 + x_2^2 + …. + x_9^2 = 720$ | 9 | 9, 12, 11, 16, 2, 5, 2, 2, 9 |
| 9 | $x_1^2 + x_2^2 + …. + x_{10}^2 = 956$ | 10 | 12, 4, 14, 8, 13, 3, 11, 13, 8, 2 |
| 10 | $x_1^2 + x_2^2 + …. + x_{11}^2 = 1502$ | 11 | 23, 7, 5, 18, 14, 10, 9, 6, 7, 8, 7 |
| 11 | $x_1^2 + x_2^2 + …. + x_{12}^2 = 3842$ | 12 | 26, 4, 14, 3, 16, 14, 43, 17, 11, 11, 8, 7 |

**Table 3:** *Results on equations with different variables*

### 4.2 Results on equations with varying number of variables

Table 3 shows the results obtained when the system was run on Diophantine equations with varying number of variables. The table consists of a list of equations of representative nature. Here the equations have the number of variables change from 2 to 12. As the table shows, the system provides solutions even when the number of variables is competitively high. The solutions found show the nature of solution offered by the procedure. The coordinates of the solution are varied, different and not closely placed. The number of generations used to find the solution also reveals the random nature of the procedure. There is a connection between the number of variables of the equations and the number of generations required to find solutions. If the number of variables is less the system gives the solutions much faster as discussed in the next section.

### 4.3 Comparing graphs of number of iterations on degree and no of variables.

Figures 3 and 4 show the relation between generations required to find solution of a given Diophantine equation and the type of the equation. Figure 3 conveys the same between the number of variables in the first nine equations given in the table 3 and the number of generations used by the system to find the first solution. It shows that if the number of variables is less the system provides the solution much faster. As the number of variables increases the generations required and hence the time taken to provide a solution also increases. As the number of variables in an equation increases, the search space also increases and hence the complexity of the search becomes large. Hence, finding of the solution of a given Diophantine equation becomes a slow process.

However, there is no such relation between the number of generations and the powers used in the equations. Figure 4 shows the effect of different of powers in finding the solution of first nine Diophantine equations given in the table 2. The graph shows that there is no apparent connection between the value of the power and the number of generations required. It shows that increase in the maximum power of an equation does not slow down the optimization process. However, in some situations, the procedure could offer the solution much faster than we expected. The powers of an equation do not affect the convergence process because as the powers increase, the search space does not increase substantially on comparison with that of the number of variables. As is the case here, the corresponding increase in values of 'N' does not affect the procedure in the faster offering of solution.

Thus, the figure 3 and figure 4 show the slightly different effect of the number of variables and powers in the finding of solutions. On comparision, as the number of variables have much greater say in slowing down the optimization processs than the powers of an equation. It has been observed that as the number of variables increase, the iterations required to find a solution increase tremendously. This is because,as the number of variables increases, there is an expoential increase in the number of possible solutions in the search space.

### 4.4 Results on equations with varying number of variables and powers

Table 3 shows the results obtained for different equations which have both powers and number of variables changed. These elementary equations are representative in nature and show the tendency of the system to deal with such the powers in the range of 2 and 5. The results show the similar trend shown by the system in the earlier cases. These equations have non-identity coefficients unlike the equations given in table 2 and table 3. The equations have number of variables change from 3 to 5 and powers do not have much impact in slowing the system in giving the solution. However, increasing number of variables has a deeper impact on the delay in producing the solution. If the number of variables is large, the system takes lots of time in giving solution. This is because, as the number of variables increases, the number of elements in the search space increases exponentially slowing down the search substantially.

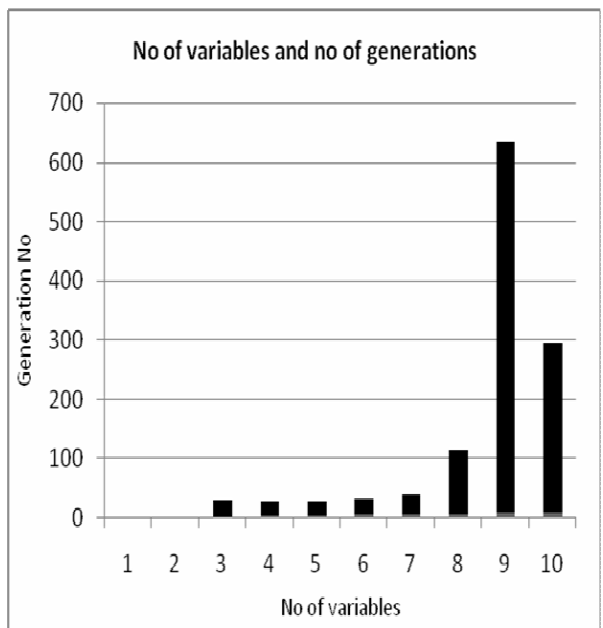

**Figure 3:** *No of variables and generation*

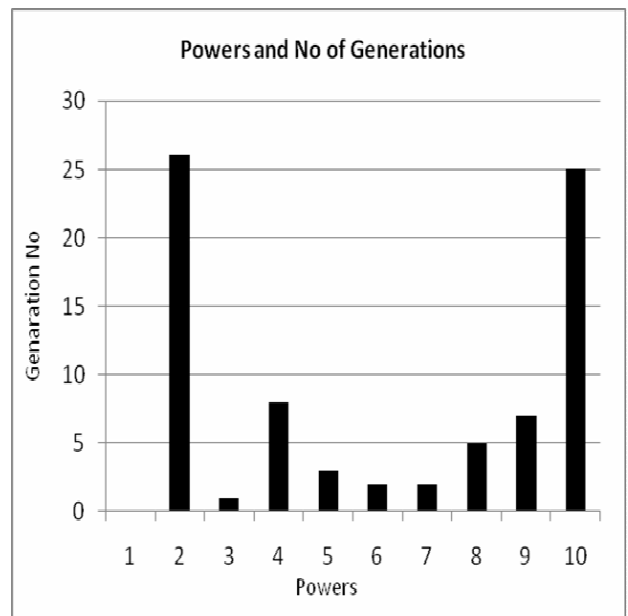

**Figure 4:** *Powers and generations*

| Sr. No | Diophantine Equation | Solution Found |
|---|---|---|
| 1 | $2x_1^2 + 6x_2^3 + x_3^2 = 1825$ | 3,40,6 |
| 2 | $3x_1^3 + 13x_2^2 + 10x_3^3 = 25632$ | 12,24,36 |
| 3 | $17x_1^3 + 12x_2^2 + 2x_3^3 = 59050$ | 10,30,25 |
| 4 | $8x_1^4 + 12x_2^2 + 2x_3^3 + 7x_3^3 = 97800$ | 10,20,30,40 |
| 5 | $5x_1^5 + 9x_2^3 + 3x_3^2 + 7x_4^2 + 2x_5^2 = 60500$ | 5,15,25,35,45 |

**Table 3:** *Results on equations with different variables and powers*

Figure 5 shows the number of generations required by these equations to provide the first solutions. Here, the x-axis represents the serial number of the equations given in table 3 and y-axis represents the number of generations required to find the first solution. The figure conveys that these equations take larger number of iterations to find solutions on comparing with equations where the powers or the number of variables alone change. This comparatively larger number of generations is due to the fact that as both powers and variables change, there will be large variations of the fitness values of the particles. This variation is supported by the non-identity coefficients in the equations. These coefficients support the fitness value of the concerned particle change much significantly. Even slight change in the value of the coordinate has a larger impact on the fitness value. As the fitness values fluctuate over a larger range, the process takes time to be steady and mature. When the procedure becomes stabilized, the procedure is more directed towards giving solutions.

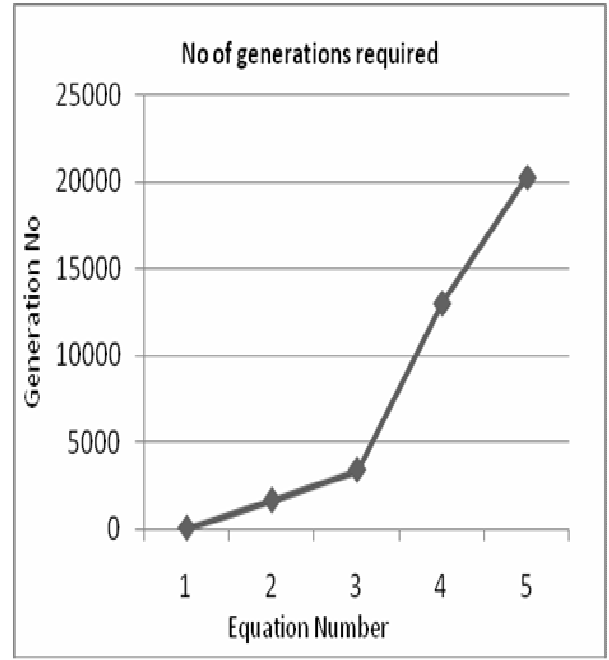

**Figure 5:** *No of generations*

As the variations become more predominant in the initial stage, the system takes more effort and time to settle the procedure to direct towards a solution. In this process, as discussed, the existence of coefficients having values other one in the equations still reduce the speed of convergence of the process. This is because larger values of coefficients will have greater role in the fluctuations of the fitness values and hence takes more time to settle down. Though, the system takes little more time to provide the solutions, the figure suggests that simultaneous influence of variables, powers and non-identity coefficients can only slightly delay the process of a finding solution.

### 4.5 Convergence of pbest Values

Figure 6 shows the convergence of the averages of the fitness values of the 'pbest' particles for the equation $x_1^2 + x_2^2 = 4500$. Initially, the averages reduce rapidly and then settle slowly. Finally, the convergence process gives the solution (60, 30) at the 29[th] generation. This is typical for any convergences procedure of 'pbest' averages. The rapid reduction at the initial stage of the optimization is due to realization of better 'pbest' values initially. Once the process becomes matured and steady, the 'pbest' values become consistent for a longer time and don't change very fast. The process takes a very slow and steady process when it is nearing the solution. Here, the fitness values are not susceptible to high fluctuations. The better consistency of the fitness values are also due to the clamp the system puts on the velocity to have a better local search of the search space. In general, the particles follow a rapid and sudden convergence at the initial stage of the convergence procedure and then settles down to a much slower and steady process before finding the solution of the given Diophantine equation.

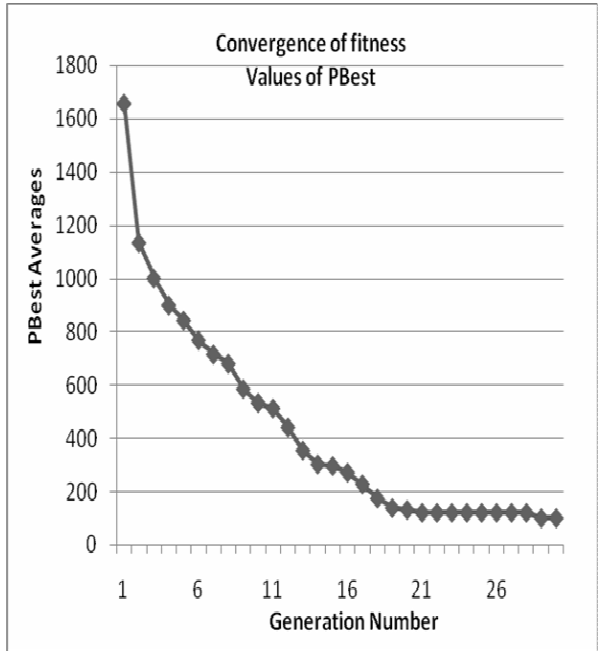

Figure 6: *Convergence of Pbest Averages*

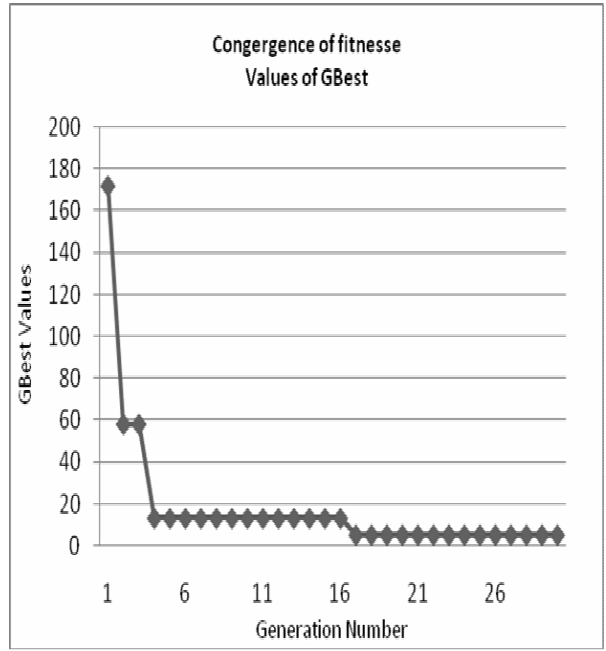

Figure 7: *Convergence of Gbest Values*

### 4.6 Convergence of gbest Values

Figure 7 shows the convergence of the fitness values of the 'gbest' particles. The example considered is the same which was used previously i.e. $x_1^2 + x_2^2 = 4500$. As the graph shows the convergence is pronounced and sudden at the initial stage of the process on comparing with the convergence of fitness averages of 'pbest' particles. Since there are much faster and drastic changes in the fitness values of the particles at the initial stage of the convergence, the convergence of 'gbest' becomes very rapid. Once the process becomes matured and steady, the convergence becomes slow. In a matured run, there would not be large fluctuations in the fitness values of the particles. Hence, 'gbest' values also do not change drastically. This reduces the convergence speed of the process considerably.

Sometimes, the 'gbest' might survive for a longer time. This is because the attainment of highly superior 'gbest' particle at that stage of the convergence. The presence of highly superior 'gbest' value for a longer time is not ideal for a good convergence. This, along with presence of same 'pbest' particles will not be in a position to produce fresh and diverse positions for the particles. The procedure has a better chance to settle down to a matured run if there is a dynamic movement of particles to fresher and diverse positions. This would help in finding the solutions faster.

### 4.7 Occurrence of gbest particles

Figure 8 shows the 'gbest' particles created by the SWARM-DOES procedure when the experiment was run for the equation $x_1^2 + x_2^2 = 10125$. Nine different 'gbest' positions of the particle where created, which were used in the subsequent generations to find the positions of the particles. The graph shows the occurrence of the 'gbest' particles in the search of the given Diophantine equation. The occurrence of these particles are scattered throughout the search space. The location of such nodes in the search space demonstrates the directed random nature of the SWARM-DOES methodology. During the process of finding numerical solution to the given Diophantine equation, the procedure could use these 'gbest' values effectively to manoeuvre the search space successfully. The number of 'gbest' particles encountered depends on the type and nature of the equation. There is no direct relation between the number of 'gbest' particles created and the solution obtained. The quality of the 'gbest' is important than the number of such particles.

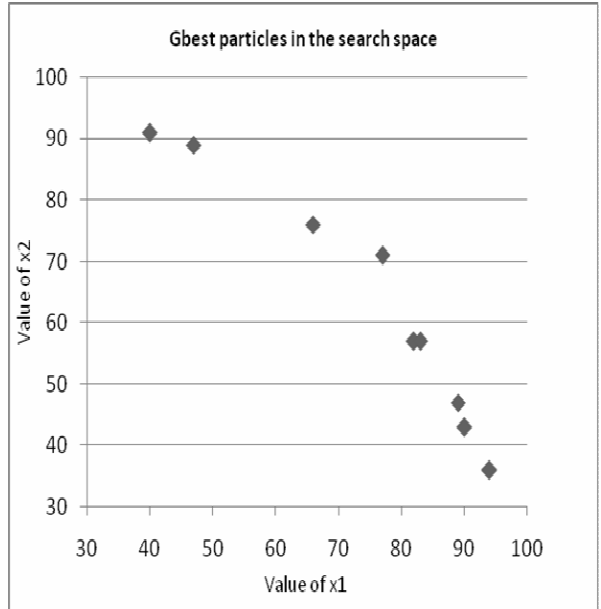

Figure 8: *Occurrence of gbest particles in the search space*

### 4.8 Cardinality of gbest particles

Figure 9 shows cardinality of 'gbest' particles created in the previous section. The cardinality of a 'gbest' particle is defined as the number of consecutive generations where a 'gbest' particle repeated in the process of finding numerical solution of an equation. The distinct numbers of repetitions of the 'gbest' particles show the random nature of the procedure in SWARM-DOES. There is no way one can predict the occurrence of a 'gbest' particle or the number of generations the same is going to survive during the optimization process. In addition, there is no apparent relationship between the number of 'gbest' particles or its number in the creation of the solution. Sometimes, a single occurrence of 'gbest' particle can produce a solution for the equation. If the procedure is successful in finding a very good 'gbest' particle during the initial stages of the optimization, there is greater chance of its survival rate for more a long time.

The successive generation of new and effective 'gbest' particles is a hallmark of a highly dynamic search process. The effectiveness of a 'gbest' particle depends on the creation of the better positions for the particles flowing in the multidimensional socio-cognitive space of the possible solutions. Though, better 'gbest' particle has a very positive influence in guiding to find better positions for the particles in the process, it comes with some shortcomings. If 'gbest' particle survives for a very long time in the optimization process due to its sheer superior nature of its fitness value, it will have a negative impact in the selection of positions of other participating particles. Fresher and diverse positions cannot be guaranteed in such situations. Only better and dynamic 'gbest' and 'pbest' particles can only guarantee an effective search for finding a solution of a Diophantine equation

### 4.9 Update of velocity of particles

Figure 10 shows the quantity of update of velocity of particles in a population. The result is corresponding to the equation $x_1^3 + x_2^3 = 16625$. This shows the random nature of the update of velocity of each particle. The figure shows that such updates takes place on a continuous basis. This is terminated only when the system settles down to the solution (25, 10) at the 40$^{th}$ generation. As the equation (10) shows, such updates depend on three factors: (i) the present velocity of the particle; (ii) the factor which is influenced by the velocity of the 'pbest' and (iii) the factor, which is influenced by the velocity of 'gbest' of the particle. The updates within a smaller range of (0, 4) is due to the restriction the system imposed on the velocity changes, which was facilitated using $v_{max}$. The presence of $v_{max}$ restricts the update of larger values and hence the large variation in the positions of the particles. Such a clamp on velocity provides better and thorough local search in the process of finding numerical solution of a Diophantine equation. If the $v_{max}$ was not available, the velocity values would have taken larger updates and hence particle might take positions, by skipping the solution position in the process.

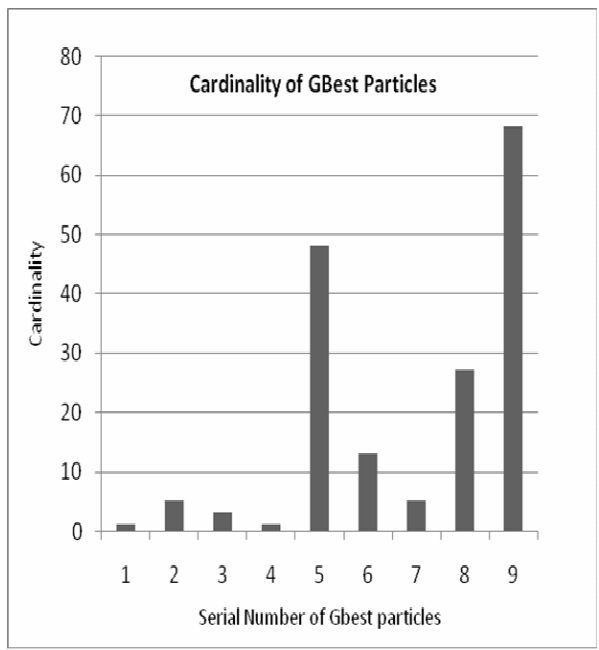

**Figure 9:** *Cardinality of gbest particles*

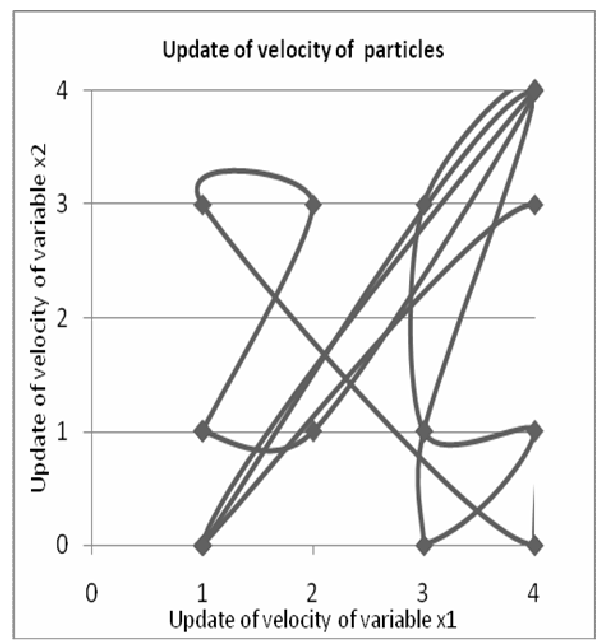

**Figure 10:** *Update of velocity of particles*

### 4.10 Statistical measures of Fitness function

Figure 11 shows different statistical measures generated corresponding to the equation $x_1^2 + x_2^2 = 10000$. The statistical measures used are mean, median and standard deviation. The measures are based on the fitness values of the particles in the SWARM-DOES process. These measures reveal the characteristics of the optimization process undertaken in the procedure. The graph shows that mean and median show a random and fluctuating nature of the fitness function values encountered in the procedure. Initially, there is much deeper change in these measures due to better positions encountered by particles. Once the process becomes stabilized and mature, such drastic changes become less. Both the measures mean and median show similar tendency as the figure shows. It reveals that the process has a stabilization capacity inbuilt in the procedure. The standard deviation lies within a small range showing the effectiveness of the procedure to direct towards the solution even when randomness plays its role. It shows the lesser variability of the fitness values of the particles with respect to its mean value during the process of finding solution to a Diophantine equation. These measures as a whole demonstrate the inbuilt strength and capacity to steer the complex search space quite effectively.

### 4.11 Effect of population size

Figure 12 shows the effect of population size on speeding up the optimization process. The figure shows the results obtained for the equation $x_1^2 + x_2^2 = 2600$. The experiments were conducted for the equation with different population size as mentioned. The number of particles taken in the experiment varied from 10 to 100 with an interval of 10. If the population size is very small, the number of generation required for finding a solution is comparatively large. Any population size in the vicinity of around 30 to 50 is sufficiently good. After that, just by increasing the population size does not give much better result. Though the results were shown for an elementary equation, similar results were obtained for other equations also.

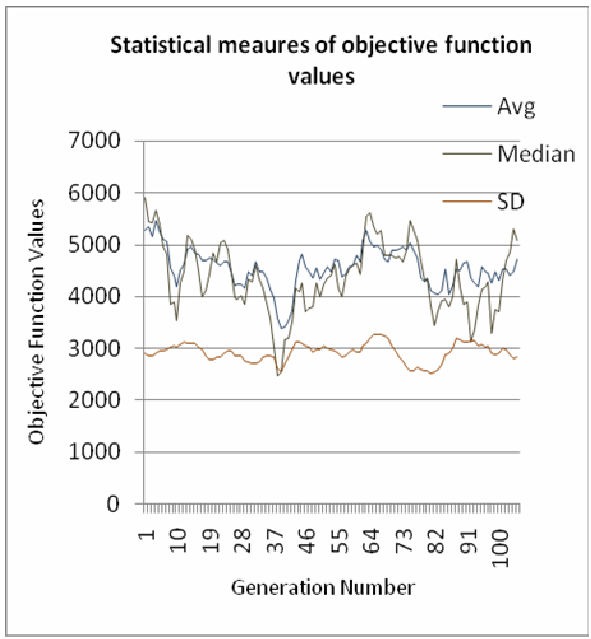

**Figure 11:** *Measures of values of fitness function*

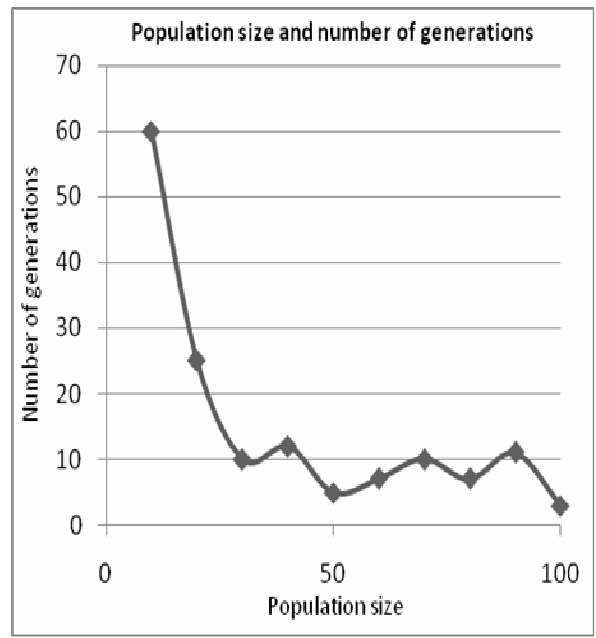

**Figure 12:** *Effect of population size*

### 4.12 Comparison with other techniques

Table 5 gives the comparative study of the results obtained by SWARM DOES methodology with other techniques. The techniques covered in the discussion are BFS, DFS, Steepest Ascent Hill Climbing, A* Algorithm, Genetic algorithm and SWARM DOES. The comparative study has been done on parameters which are common to the techniques discussed.

- **Representation of solutions**: The candidate solutions are represented as vectors in all methods except on GA and SWARM DOES where they are represented as chromosomes (or strings) and particles respectively. The proper representation helps to find solution of the equation in each of the case. The vector representation comes quite natural to the first four methods as the same is in sync with the requirement of these algorithms. The string representation in GA helps to have the meeting of the proper requirements of the particular equation along with that of the genetic algorithm. The particles in the SWARM DOES system are represented as arrays.

- **Computation Style**: The computation style followed in the methods is sequential except on GA and SWARM-DOES where the style is parallel. The parallel computation style of the GA and SWARM-DOES works effectively while dealing with huge number of candidate solutions in the search space. In the case of GA, a population of strings are in search of solutions concurrently while in the case of SWARM DOES, a population of swarm perform at the search. This parallel computing framework of chromosomes or particles help to find the solutions of the equation much more effectively than other algorithms discussed in the work.
- **Procedure type:** The procedure followed in all the techniques except GA and SWARM DOES are deterministic where they follow randomly. Though, the sequential way of moving towards the solution have many advantages, especially when the equations are elementary and easy to manage, often they bring lot of computational complexity. These methods are not guaranteed to give solutions when the equations are quite large and the value of 'N' is very high as the number of candidate solutions in the search space is extremely large. The directed random nature of GA and SWARM-DOES are effective in managing the complexity of search space.
- **Mathematical structure**: The mathematical structure followed in BFS, DFS, A* and Hill Climbing are tree structure. Nodes are generated as and when a new candidate state is searched and finding solution is translated to traversing through the tree. While GA does not have a formal mathematical structure, SWARM DOES follows a proper structure. It uses fully connected graph as the neighbourhood topology. Through this, the procedure maintains a relation between each and every particles participating in the optimization process. It helps to keep track of each particle and link with its best performance, which is incorporated in the procedure for optimum results. The linking with the all the particles in the population, not just a couple of neighbours, help to fine tune each particle's search mechanism in a way that is in tune with the best performance of other particles.
- **New states formed:** The new candidate solutions are formed using typical production rules in the first four techniques where as GA and SWARMDOES follow stochastically. The stochastic nature of creation of new states helps in reducing the complexity of search space.

| *Characteristics.* | *BFS* | *DFS* | *A\** | *Hill Climbing* | *GA* | *SWARM DOES* |
|---|---|---|---|---|---|---|
| *Representation of solution* | Vector | Vector | Vector | Vector | Chromosomes | Particles |
| *Computational style* | Sequential | Sequential | Sequential | Sequential | Parallel | Parallel |
| *Procedure type* | Deterministic | Deterministic | Deterministic | Deterministic | Random | Random |
| *Mathematical structure* | Tree | Tree | Tree | Tree | NA | Graph |
| *New state formed* | Using production rules | Using production rules | Using production rules | Using production rules | Stochastically | Stochastically |
| *Occurrence of Local Optima* | No | Yes | No | Yes | Yes | No |
| *Local optima Tackled* | NA | Backtracking | NA | Backtracking | By Coevolution | NA |
| *Convergence to solution* | Slow | Slow | Rapid, then steady | Rapid, then steady | Rapid, then steady | Rapid, then steady |
| *Appearance of Solution* | Node | Node | Node | Node | Evolved-chrom | Evolved-particle |
| *Coordinates of solution* | Close | Close | Close | Close | Distinct | Distinct |

***Table 5**: Comparative study*

- **Occurrence of local optima**: The high light of the SWARM DOES methodology with respect to other methods is the effectiveness in tackling the occurrence of local optima by SWARMDOES. This does encounter any local optima the search process.
- **Local Optima Tackled**: SWARM-DOES do not come across local optimum points. Most of the other methods require some additional procedure needed to tackle such optimum points encountered during the search procedure.
- **Convergence to solution**: The process of convergence to the solution shows a certain trend. BFS and DFS follow a very slow convergence whereas other techniques initially follows a rapid convergence, then stabilises and move through a steady momentum before converging to solution.
- **Appearance of solution**: The initial four methods show the node as the solutions whereas GA shows the chromosome and SWARM DOES shows the solution as evolved particle.
- **Coordinates of the solution**: the coordinates of the solution show the effectiveness of the technique under study. When the first four techniques give solution the coordinates are closely placed. They fail to give distinct looking solution. The GA and SWARM DOES could give solutions whose coordinates are distinct and different looking.

Though, GA and SWARM DOES could give solutions on better way than other techniques, SWARM DOES is better positioned to find solutions of a Diophantine equation because of its sheer effectiveness in handling larger and complex equations.

## 5 CONCLUSION AND FUTURE WORK

The paper describes a procedure which uses particle swarm optimization as a methodology for finding numerical solutions of a Diophantine equation, which does not have a formal method to find solution. The particles are represented as integer particles and are guided using domain specific fitness function. The particle best and global best positions of the particles help the procedure to move towards the solution. The procedure follow fully connected neighborhood topology. The experimental results showed that procedure could offer solutions having distinct coordinates. The method performs effectively with very large values of powers and N and reasonable number of variables.

The further works involve in scaling the procedure in tackling very large equations having very large number of variables. The authors would like to increase the efficiency of the system by incorporating domain specific heuristic to deal with such complexity. It is expected that such measures would help to find solutions of vary large equations with large number of variables.

## WEBSITES

http://www.swarmintelligence.org
http://www.softcomputing.net
http://www.mauriceclerc.net.
http://www.wikipedia.org
http://www.scholarpedia.org
http://www.birs.ca/workshops/2007/07w5063/report07w5063